\pdfoutput=1

\documentclass[11pt]{article}

\usepackage[preprint]{acl}
\usepackage{amsmath}
\usepackage{algorithm,enumitem}
\usepackage[noend]{algpseudocode}
\usepackage{algorithmicx}

\usepackage{times}
\usepackage{latexsym}
\usepackage[T1]{fontenc}

\usepackage[utf8]{inputenc}

\usepackage{microtype}
\usepackage{epsdice}

\usepackage{inconsolata}

\usepackage{graphicx}

\title{Agentic Verification for Ambiguous Query Disambiguation}

\author{\textbf{
Youngwon Lee\thanks{~Work done while visiting Snowflake. Correspondence to:  \href{mailto:seungwonh@snu.ac.kr}{\texttt{seungwonh@snu.ac.kr}}.}\quad
Seung-won Hwang\footnotemark[1]\quad
Ruofan Wu\textsuperscript{\textdagger}\quad
Feng Yan\textsuperscript{\textdagger}}\\
\textbf{
Danmei Xu\quad
Moutasem Akkad\quad
Zhewei Yao\quad
Yuxiong He
}\\
Snowflake AI Research\quad
{}\footnotemark[1]Seoul National University\quad
{}\textsuperscript{\textdagger}University of Houston\\
}

\usepackage{comment}
\usepackage{amsmath}
\usepackage{pifont}
\usepackage{subcaption}
\usepackage{caption, booktabs}
\usepackage{amssymb}
\usepackage{setspace}
\usepackage{bbm}
\usepackage{afterpage}
\usepackage{enumitem}

\usepackage{xparse}

\algrenewcommand\algorithmicrequire{\textbf{Input:}}
\algrenewcommand\algorithmicensure{\textbf{Output:}}

\makeatletter
\newcommand\footnoteref[1]{\protected@xdef\@thefnmark{\ref{#1}}\@footnotemark}
\makeatother
\newcolumntype{P}[1]{>{\centering\arraybackslash}p{#1}}

\usepackage{xspace}
\newcommand{\ours}[0]{\textsc{VerDICt}\xspace}
\newcommand{\ourslong}[0]{Verified-Diversification with Consolidation\xspace}

\usepackage{adjustbox}
\makeatletter
\newcommand{\thickhline}{
    \noalign {\ifnum 0=`}\fi \hrule height 1pt
    \futurelet \reserved@a \@xhline
}
\newcolumntype{"}{@{\hskip\tabcolsep\vrule width 1pt\hskip\tabcolsep}}
\makeatother
\usepackage{amsfonts}

\makeatletter
\newcommand*{\blackleq}{
  \mathrel{
    \mathpalette\@blackleq{}
  }
}
\newcommand*{\@blackleq}[2]{
  \vcenter{
    \m@th
    \setbox0=\hbox{$#1\mkern3mu$}
    \setbox2=\hbox{$#1\vcenter{}$}
    \setbox4=\hbox{\raisebox{-\ht2}[.2pt][.2pt]{$#1-$}}
    \hbox{$#1\blacktriangleleft$}
    \nointerlineskip
    \kern\wd0 
    \copy4 
  }
}
\makeatother
\usepackage{dsfont}
\newcommand{\argmax}{\operatornamewithlimits{argmax\text{ }}}

\usepackage{multirow}

\usepackage{mathtools}
\newcommand{\argtopk}{\operatornamewithlimits{arg\text{ }topk\text{ }}}

\usepackage{placeins} %
\usepackage{tcolorbox}

\begin{document}
\maketitle

\begin{abstract}

In this work, we tackle the challenge of disambiguating queries in retrieval-augmented generation (RAG) to diverse yet answerable interpretations.
State-of-the-arts follow a Diversify-then-Verify (DtV) pipeline, where diverse interpretations are generated by an LLM,
later used as search queries to retrieve supporting passages.
Such a process
may introduce noise in either interpretations or retrieval,
particularly in enterprise settings, where LLMs---trained on static data---may struggle with domain-specific disambiguations.
Thus, a post-hoc verification phase is introduced to prune noises.
Our distinction is \textbf{to unify diversification with verification} by incorporating feedback from retriever and generator early on.
This joint approach improves both efficiency and robustness by reducing reliance on multiple retrieval and inference steps, which are susceptible to cascading errors.
We validate the efficiency and effectiveness of our method, \ourslong (\ours), on the widely adopted ASQA benchmark
to achieve diverse yet verifiable interpretations.
Empirical results show that \ours improves grounding-aware $\textrm{F}_1$ score by an average of 23\% over the strongest baseline across different backbone LLMs.
\end{abstract}

\section{Introduction}
\label{sec:intro}

\begin{figure*}
\centering
\includegraphics[width=.95\linewidth]{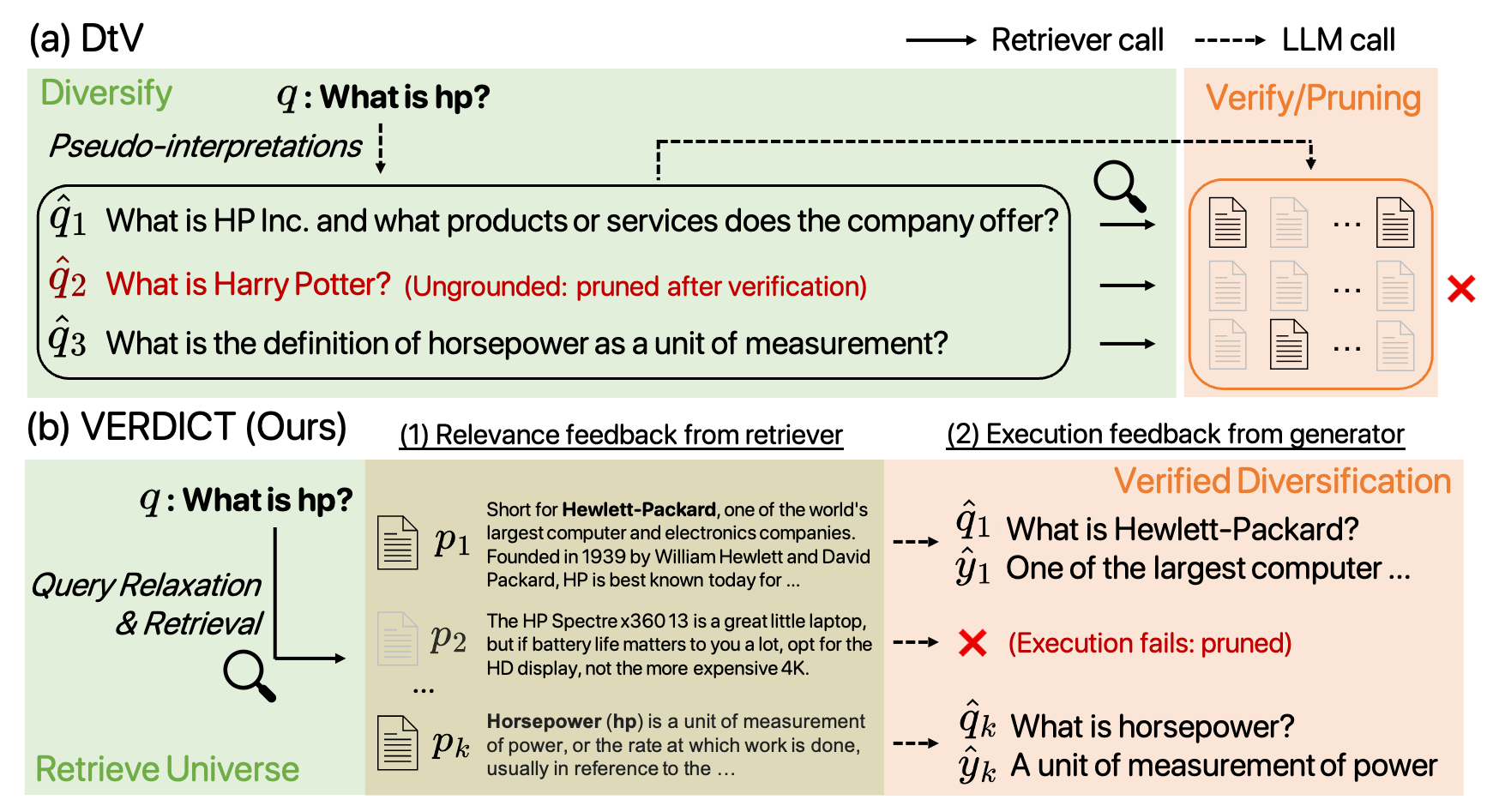}
\caption{
Comparison of (a) DtV (Diversify-then-Verify) and (b) \ours (\ourslong, ours).
}
\label{fig:overview}
\end{figure*}

Retrieval-Augmented Generation (RAG; \citealp{lewis-etal-2020-retrieval-neurips}) is expected to complement large language models (LLMs), trained on static data, when they struggle to provide reliable responses without external retrieval.
This challenge is particularly evident in enterprise settings, where queries are asked on a domain-specific and potentially evolving corpora, inaccessible during LLM training.
In such cases, RAG must (1) \textbf{disambiguate} short, vague user queries and (2) \textbf{retrieve} supporting evidence, %
to ensure the verifiability of the generated response.

A line of work on handling ambiguous question answering~\citep{min-etal-2020-ambigqa,cole-etal-2023-selectively,in-etal-2024-diversify-arxiv} adopted a Diversify-then-Verify (DtV) pipeline, which first generates diverse interpretations of the query using an LLM and then retrieves supporting documents using interpretations as search queries.
However, this process often leads to ungrounded or noisy interpretations, particularly when LLMs struggle with domain-specific or evolving meanings. For example, as shown in Figure~\ref{fig:overview}(a), the query ``What is HP'' may refer to Hewlett-Packard, horsepower, or Harry Potter.
LLMs trained on general knowledge would suggest all three, while an enterprise corpus may contain no references to Harry Potter.
In other words, this interpretation is \textbf{ungrounded}.
However, DtV still retrieves documents for all three interpretations, introducing noise from unnecessary retrieval.
Figure~\ref{fig:overview}(a) illustrates DtV, where the verification step comes after retrieval to prune out noisy passages, though 
``pseudo-interpretations'' have already  negatively influenced the process.

To address these limitations, we propose \ours (\ourslong), a unified  framework that integrates diversification and verification:
\ours is \textbf{agentic}, in the sense that we ask feedback from retriever and generator to ensure all  disambiguations are grounded upfront, as in Figure~\ref{fig:overview}(b).
This joint process, by avoiding 
noisy and unnecessary retrieval
of ungrounded interpretation,
mitigates cascading errors while improving efficiency. 
Specifically, agentic feedback from two sources play a critical role in the process:

\begin{itemize}
\item Relevance feedback from the retriever: We use a retriever to obtain a set of passages 
$U_q$, that are relevant to $q$ 
and encompass all interpretations found in the corpus. This retrieval-augmented diversification, as also employed in RAC~\citep{kim-etal-2023-tree}, naturally filters out unsupported interpretations, such as ``Harry Potter,''  as it is irrelevant, or may become obsolete as corpus evolves.

\item Execution feedback from the generator: %
Figure~\ref{fig:overview}(b) illustrates why relevance alone does not guarantee that a question can be answered, necessitating \emph{execution feedback}.
Passage $p_2$, though relevant to Hewlett-Packard as it describes its products, cannot provide an answer to the given question.
This motivates us to leverage the generator feedback to decide whether the question can be answered from the passage, based on which we prune out $p_2$.

\end{itemize}

Lastly, since feedback signals themselves can also be noisy, we introduce a consolidation phase to leverage consistency.
By aggregating  multiple feedback signals through clustering, \ours further mitigates potential noises.
This phase does not incur an additional feedback cost while enabling a more robust decision-making.

As a bonus, this inherently grounded design of \ours,
by tightly connecting each interpretation with its supporting passage, can be evaluated for
verifiability metrics popular in RAG~\citep{li-etal-2023-survey-arxiv,liu-etal-2023-evaluating}, such as citation quality.
We use both existing and grounded metrics, to show ours disambiguation is grounded and accurately answered.

Our key contributions are as follows:
\begin{itemize}
\item We introduce \ours, a novel framework that unifies diversification and verification, enabled by an
agentic approach of leveraging feedback from the retriever and the generator.
\item We demonstrate that \ours improves efficiency by reducing unnecessary retrieval and verification steps, and also effectiveness, by mitigating cascading errors common in DtV.
\item We empirically validate \ours on widely adopted ambiguous QA benchmark, ASQA~\citep{stelmakh-etal-2022-asqa}, with average gain of 23\% in $\textrm{F}_1$ score across different backbone LLMs.\footnote{The gain is computed against the best baseline for each backbone LLM.}
We also release our code and verifiability evaluation framework to facilitate future research.
\end{itemize}

\section{Related Work}

This section overviews existing work on DtV workflow and the line of research on using retriever as environment. 

\paragraph{Diversification.}
The goal of this phase is to turn the given query into a diverse set of interpretations.
\citet{min-etal-2021-joint} and \citet{sun-etal-2023-answering} studied an iterative approach by identifying one new intent at a time.
\citet{gao-etal-2021-answering} generates clarification questions conditioned on generated answers from ambiguous questions, and then answers the clarification questions again.

With LLMs,
few-shot in-context learning
was used to generate clarifications or query rewrites directly, relying solely on their internal knowledge captured during pretraining~\citep{kim-etal-2023-tree, ma-etal-2023-query}.

\paragraph{Verification.}
Errors in retrieval or intent inference may generate question-passage pair that cannot provide the expected answer.
Verification aims at pruning such pairs.
\citet{shao-huang-2022-answering} train a verifier to choose correct answers from a list of candidates drafted from each passage without question clarifications.
More recent works such as 
Self-RAG~\citep{asai-etal-2024-self-iclr} and Corrective RAG~\citep{yan-etal-2024-corrective} also train a verifier to decide whether the retrieved passages are relevant enough to assist answer generation.
To avoid verifier training, LLMs may leverage its parametric knowledge instead to verify~\citep{li-etal-2024-llatrieval}.

\paragraph{Relevance Feedback.}
In information retrieval, 
imperfect user queries are often modified, guided by treating
initial retrieval result~\citep{rocchio71relevance}
as pseudo relevance feedback (PRF)~\citep{efthimiadis-biron-1993-ucla-trec,evans-lefferts-1993-design-trec,buckley-etal-1994-automatic-trec}.
In our problem context, retrieval-augmented clarification in RAC~\citep{kim-etal-2023-tree} and our method, can be interpreted as leveraging PRF to extract diversified interpretations;
we provide more detailed comparison in Section~\ref{subsubsec:baselines}.

\paragraph{Our Distinction.} 

Unlike DtV that considers verification as a post-hoc step,
we jointly pursue diversification and verification.
Unlike RAC leveraging relevance feedback only,
we jointly verify with execution feedback,
followed by consolidation of feedback for further denoising.

\section{Preliminary}

\subsection{Problem Formulation}
\label{sec:problem_formuation}

Our formulation adheres to the conventional ambiguous question answering setting, where the primary objective is to handle ambiguity in user queries effectively. Specifically, given an ambiguous question \( q \), the task of diversification involves identifying a comprehensive set of possible interpretations \( \mathcal{Q} = \{q_1, q_2, \cdots, q_N\} \), each representing a distinct and valid meaning of the original question $q$. 

The goal then extends to determining corresponding answers \( \mathcal{Y} = \{ y_1, y_2, \cdots, y_N \} \) for each interpretation \( q_i \), ensuring the response is grounded on retrieved supporting evidence.
The established evaluation protocol for such task involves comparing the model-generated predictions \( \hat{\mathcal{Q}} \) and \( \hat{\mathcal{Y}} \) against the ground-truth set of interpretations and answers, \( \mathcal{Q} \) and \( \mathcal{Y} \).

\subsection{Baselines: DtV}
\label{subsec:diva}

In this section, we formally describe 
the DtV workflow, using the most recent work \citet{in-etal-2024-diversify-arxiv} with state-of-the-art performance as reference.
DtV first \textbf{diversifies} the query, or, identifies pseudo-interpretations \( \hat{\mathcal{Q}} = \{\hat{q}_1, \hat{q}_2, \dots, \hat{q}_M\} \) that disambiguate the meaning of the original ambiguous query, by prompting LLM with instructions $I_{\textrm{P}}$ without accessing any retrieved knowledge:
\begin{equation}
\hat{\mathcal{Q}} \leftarrow \texttt{LLM}(q 
; I_\text{P}).
\label{eq:diva_q_extract}
\end{equation}
Each generated pseudo-interpretation $\hat{q}_i$ in $\hat{\mathcal{Q}}$ are then used as a search query to retrieve top-$k$ supporting
passages, the union of which
forms a universe $U$
of relevant documents
\begin{equation}
U_{\hat{\mathcal{Q}}} \leftarrow \bigcup_{\hat{q}\in\hat{\mathcal{Q}}}\, \argtopk_{p} \mathrm{sim} \left( \hat{q}, p \right),
\label{eq:diva_form_universe}
\end{equation}
where the subscript $\hat{\mathcal{Q}}$ indicates that the universe $U_{\hat{\mathcal{Q}}}$ is derived from the set of pseudo-interpretations $\hat{\mathcal{Q}}$.
As pseudo-interpretations can be ungrounded and the resulting $U_{\hat{\mathcal{Q}}}$ can be noisy,
\textbf{verification} phase follows, to examine each pseudo-interpretation 
with universe $U_{\hat{\mathcal{Q}}}$.
After this phase,
$U_{\hat{\mathcal{Q}}}$ is reduced to a verified subset 
$U_V$,
from which disambiguated queries are answered.
\begin{equation}
U_V
\leftarrow \left\{
p\in U_{\hat{\mathcal{Q}}}
\,\middle|\, 
\exists_{\hat{q} \in \hat{\mathcal{Q}}}\,
\texttt{Verify}(\hat{q}, p)=1 \right\},
\end{equation}
\begin{equation}
\hat{\mathcal{Q}}^\prime, \hat{\mathcal{Y}} \leftarrow 
\texttt{LLM}(q, U_{V}; I_{\textrm{G}}).
\label{eq:diva_answer_gen}
\end{equation}

We identify the challenges and inefficiencies in DtV line of work as follows:
\begin{enumerate}

\item $U_{\hat{\mathcal{Q}}}$:
Each interpretation $\hat{\mathcal{Q}}$ incurs at least one retriever call, some of which are ungrounded and could introduce noises.

\item $U_{V}$:
Verified $U_V$ requires to process 
all $(\hat{q},p)$ pairs, for example, inducing $|\hat{\mathcal{Q}}|$ \texttt{Verify} calls with input size of $\mathcal{O}(|U_{\hat{\mathcal{Q}}}|)$ each.\footnote{\emph{Pruning} $U_{\hat{\mathcal{Q}}}$ in advance
 was used in~\citet{in-etal-2024-diversify-arxiv}, though asymptotic cost remains unchanged. We discuss this in Section~\ref{subsec:clustering}.}
Plus, such long-context verification requires a powerful LLM model,
which increases cost and hinders applicability.

\end{enumerate}

\begin{figure*}[tbp]
\centering
\includegraphics[width=\linewidth]{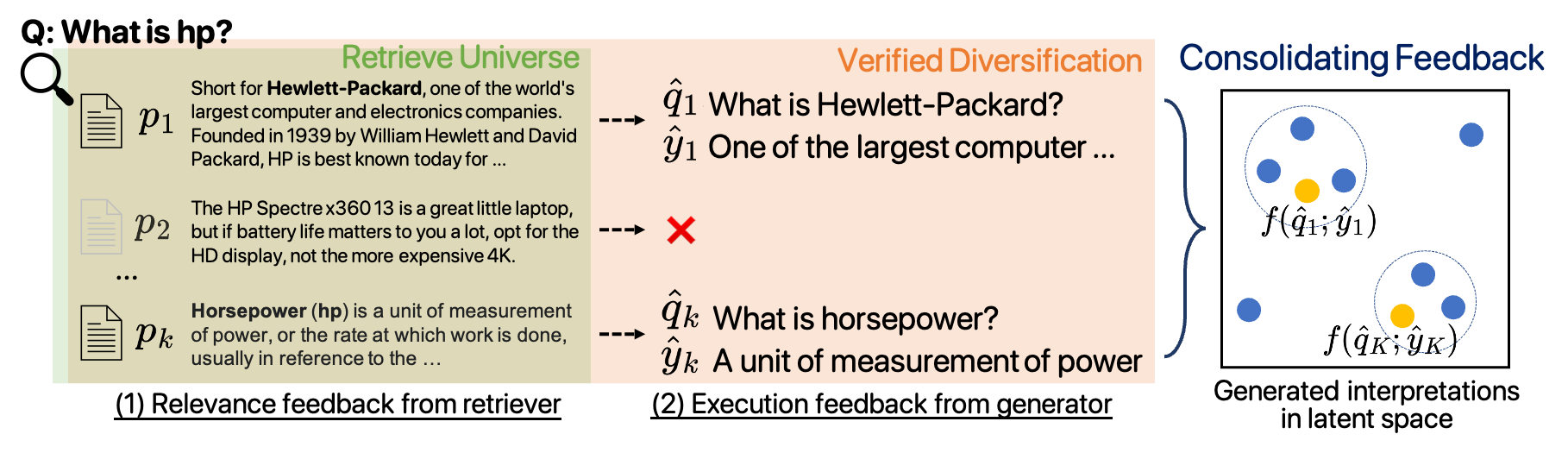}
\caption{
Illustration of the full pipeline of \ours: Verified diversification (Section~\ref{subsec:extraction}) followed by consolidation phase (Section~\ref{subsec:clustering}).
On the right, yellow and blue dots represent embeddings of generated interpretations and their answers, embedded together after concatenation, while yellow color denotes medoids, or representatives chosen from each cluster.
}
\label{fig:ours}
\end{figure*}

\section{Method}
\label{sec:method}

We illustrate \ours with two distinctions:
Verified Diversification (Section~\ref{subsec:extraction})
and 
Consolidating Feedback (Section~\ref{subsec:clustering}).
In Section~\ref{subsec:eval}, we provide grounded evaluation protocols.

\subsection{Verified Diversification}
\label{subsec:extraction}

Unlike DtV,
we ensure that $\hat{\mathcal{Q}}$ includes only grounded interpretations by leveraging agentic feedback from both the retriever and the generator.
Unlike Eq.~\ref{eq:diva_q_extract}, where pseudo-interpretations are conditioned solely on the ambiguous query $q$, our approach conditions on both the query $q$ and the corpus $C$.
This ensures that each interpretation $\hat{q} \in \hat{\mathcal{Q}}$ is explicitly grounded to at least one passage from the corpus, ensuring 
both \textbf{relevance} to $q$, using the relevance feedback of retrieving $U_q$, and \textbf{answerability} of the interpretation.

\begin{equation}
\hat{\mathcal{Q}},
\hat{\mathcal{P}},
\hat{\mathcal{Y}} \leftarrow 
\textrm{VD}(q,U_q).
\label{eq:ours_q_extract}
\end{equation}

Note that Eq.~\ref{eq:ours_q_extract}
jointly infers
whether the retriever can retrieve a relevant passage (relevance) and an answer to $q$ can be generated from the retrieved passage.
As a result,
interpretations $\hat{\mathcal{Q}}$ and answers $\hat{\mathcal{Y}}$ are obtained simultaneously, along with the list of passages $\hat{\mathcal{P}}$ that support each identified question-answer pair.

\paragraph{Relevance Feedback.}

``Relevance feedback from the retriever'' in Figure~\ref{fig:ours} shows how
we first retrieve
$U_q$ with a single round of retrieval with the original query $q$ before generating interpretations. To ensure high recall within the universe, we use $q'$ relaxed as below:

\begin{equation}
q^\prime \leftarrow \texttt{LLM}(q; I_{\textrm{R}}),  
\end{equation}
\begin{equation}
U_q \leftarrow \argtopk_{p} \mathrm{sim} \left( q^\prime, p \right).
\end{equation}
With relaxed $q^\prime$ and
choice of a larger \(k\) for retrieval,\footnote{$k$ was empirically tuned to 20, which nearly matched G-$\textrm{F}_1$ (described in \S~\ref{subsec:eval}) from top-100 with GPT-4o backbone.}
we further increase the diversity and coverage of the interpretations while keeping the process computationally efficient, marked as ``relevance feedback from the retriever'' in Figure~\ref{fig:ours}.

\paragraph{Execution Feedback.} 
Execution feedback from the LLM generator in Figure~\ref{fig:ours}
prunes interpretatons that cannot be answered.
Conditioned on our high-coverage universe $U_q$, the LLM generator is prompted to identify interpretations $\hat{\mathcal{Q}}$ and corresponding answers $\hat{\mathcal{Y}}$ from \emph{each} passage.

This ensures the answerability of each disambiguated query $\hat{q}_i$, given the passage $\hat{p}_i$ from which it was derived:
For example, in Figure~\ref{fig:ours},
passages like $p_2$, which mentions HP products but cannot answer the query asking what HP is--are effectively pruned through failed execution of generating question-answer pair.
In contrast, with passages like $p_1$, a valid question-answer pair is generated.
Such execution feedback is formalized as follows,
\begin{equation}
(\hat{q}_i, \hat{y}_i) \leftarrow \texttt{LLM}(q, p_i; I_{\textrm{E}}),
\label{eq:ours_q_extract_single}
\end{equation}
of which the LLM prompt $I_{\textrm{E}}$ can be found in Figure~\ref{fig:prompt_q_extract}.

From a practical standpoint, Eq.~\ref{eq:ours_q_extract_single} can be executed in parallel, with each LLM call processing a single passage at a time.
This minimizes the input sequence length for each call, optimizing both latency and computational overhead while reducing hallucination in less capable models.

\begin{algorithm*}[t]
\caption{Verified Diversification for ambiguous queries}
\begin{algorithmic}[1]
\Require Question $q$,
LLM $\texttt{LLM}(\cdot)$
\Ensure Pairs of clarification question and answer $\hat{\mathcal{Q}}=\{(\hat{q}, \hat{y})\}$

\State $q^\prime$ $\leftarrow$ Relax $q$ for high-recall universe
\State %
$U$ $\leftarrow$ Top-$k$ retrieved passages from the retriever, using $q^\prime$ as query

\State $\mathcal{Q} \leftarrow \{\}$
\For{$i=1$ to $k$}
    \State $(\hat{q}_i, \hat{y}_i)$ $\leftarrow$ $\texttt{LLM}(q, p_i; I_{\textrm{E}})$ \Comment{Extracting interpretation from passage $p_i$ with execution feedback}
    \If {$\hat{q}_i$ is not \texttt{None}}
        \State $\mathcal{Q} \leftarrow \mathcal{Q} \cup \left\{ (\hat{q}_i, \hat{y}_i) \right\}$
    \EndIf
\EndFor

\State $\mathcal{C}$ $\leftarrow$ Cluster into partitions $\mathcal{C}_i$'s of $\mathcal{Q}$ based on embeddings $f(\hat{q}_i; \hat{y}_i)$

\State $\hat{\mathcal{Q}} \leftarrow \{\}$
\For{$j=1$ to $k$}
    \State $(\check{q}_j, \check{y}_j)$ $\leftarrow$ $\argmax_{({q}^*,y^*)\in \mathcal{C}_j} \sum_{(\hat{q},\hat{y})\in \mathcal{C}_j} \mathrm{sim}(f(\hat{q};\hat{y}), f(q^*;y^*))$ \Comment{Medoid of the $j$-th cluster $\mathcal{C}_j$}
    \State $\hat{\mathcal{Q}} \leftarrow \hat{\mathcal{Q}} \cup \left\{ (\check{q}_j, \check{y}_j) \right\}$
\EndFor

\State \textbf{return} $\hat{\mathcal{Q}}$
\end{algorithmic}
\label{alg:ours}
\end{algorithm*}

\begin{table}[t]
\centering
\fontsize{10.5pt}{12pt}\selectfont
\setlength{\tabcolsep}{3pt}
\begin{tabular}{lcccc}
    \thickhline
     & \multicolumn{1}{c}{\multirow{1.25}{*}{Retriever}} & \multicolumn{3}{c}{\multirow{1.25}{*}{LLM}}
     \\
    \cmidrule(r){2-2}
    \cmidrule(r){3-5}
    & \multicolumn{1}{c}{\multirow{-1.25}{*}{\# calls}} & \multicolumn{1}{c}{\multirow{-1.25}{*}{\# calls}} & & \multicolumn{1}{c}{\multirow{-1.25}{*}{input len.}} \\ \hline
    DtV       & $\mathcal{O}(|\hat{\mathcal{Q}}|)$ 
    & $\mathcal{O}(|\hat{\mathcal{Q}}|)$ & $\times$ & $\mathcal{O}(|U_{\hat{\mathcal{Q}}}|)$ \\
    \rowcolor{gray!10}
    \ours & $\mathcal{O}(1)$
    & $\mathcal{O}(|U_q|)$ & $\times$ & $\mathcal{O}(1)$ \\
    \thickhline
\end{tabular}
\setlength{\tabcolsep}{6pt}
\caption{
Cost comparison between DtV and \ours in terms of retriever and LLM calls per question. For LLM calls, input context length is given in terms of the number of passages, resulting in the total input size when multiplied by the number of calls. 
}
\label{tab:comp}
\end{table}

Table~\ref{tab:comp} provides comparison of asymptotic cost of retrieval and LLM calls in \ours and DtV.
First, while \ours
makes a single call to the retriever as both builds on $U_q$, DtV issues several search queries, proportional to the number of pseudo-interpretations $|\hat{\mathcal{Q}}|$.
Next, the number of LLM calls and input size of each call, measured in terms of the number of passages, also shows the efficiency of \ours.
DtV provides several passages as a list input, increasing the context size and the chance of introducing noise while processing the elongated input.
In contrast, 
\ours requests for answerability feedback per  each passage in $U_q$, 
using only a single passage as a  context.

\subsection{Consolidated Feedbacks}
\label{subsec:clustering}

While retriever and generator feedback are essential for verification, both are prone to noise. %
For robust verification,
we propose to consolidate retriever and generator feedback into a unified, robust signal, without additional retrieval or inference calls. %

Specifically, question-answer pair from generator  are projected into a latent space $\mathbb{R}^d$ using an encoder $f$ from the retriever (Figure~\ref{fig:ours}).
We then cluster similar pairs,
to choose those consistently supported by relevant passages, while filtering out outliers from noisy passages.
In this process,
multiple generator and retriever feedback signals are consolidated, which aligns with seeking the most consistent outputs~\citep{chen-etal-2023-codet-iclr}, improving robustness by inference scaling of aggregating multiple LLM outputs.
Additionally, selecting a medoid question per cluster deduplicates redundant interpretations, whereas DtV defers deduplication to the answer generation stage in Eq.~\ref{eq:diva_answer_gen}.

While DtV uses similarity for pruning $U$, 
we use both retriever and generator feedback, for another purpose of consolidation, to leverage richer signals including interaction between the two.

\subsection{Grounding in Evaluation}
\label{subsec:eval}

This section discusses existing  evaluation metrics, contrasted with
 grounded evaluation, often required  in enterprise RAG scenarios.

\paragraph{Existing: Ungrounded Metrics.}
Current benchmarks~\citep{min-etal-2020-ambigqa, stelmakh-etal-2022-asqa} measure whether diverse human-annotated interpretations ``match''  model-generated ones, typically using recall as the primary metric. Lexical similarity (e.g., BLEU) determines the match between generated ($\hat{q}$) and reference ($\tilde{q}$) interpretations:

\begin{equation}
V(\hat{q},\tilde{q}) \in \{0,1\}.
\label{eq:simple_match}
\end{equation}

We classify this as \textbf{ungrounded evaluation}, as it does not assess whether retrieved passage supports the generated interpretation.

\paragraph{Enterprise: Grounded Precision and Recall.}

In enterprise RAG, answer verifiability—measured through whether answer can be supported by the correct citation of a supporting passage in the corpus~\citep{li-etal-2023-survey-arxiv, liu-etal-2023-evaluating}—has been a  evaluation criterion. This concept naturally extends to verify disambiguation, where an interpretation is validated based on whether its answer can be supported by a passage in the corpus, as a binary evaluation function below:

\begin{equation}
V(\hat{q},\hat{p}) \in \{0,1\},
\end{equation}
where  $\hat{p}$
is predicted as the supporting passage associated with $\hat{q}$ by the model.

With this extended $V$,
grounded precision is defined as
the ratio of correctly grounded
among model-generated interpretations:
\begin{align}
\mathrm{G\text{-}Precision} = \frac{1}{|\hat{\mathcal{Q}}|} \sum_{\hat{q}\in \hat{\mathcal{Q}}} V(\hat{q}, \hat{p}).
\label{eq:def_precision}
\end{align}

While optimizing for existing recall may encourage ungrounded diversity, pursuing G-precision counterbalances this by assessing whether the interpretation is accurately grounded. 

Similarly, we extend  to define a grounded recall metric that evaluates model-generated interpretations,  only against the \textbf{grounded gold} set, $\bar{\mathcal{Q}}$. This ensures that generating ungrounded interpretations, such as “Harry Potter,” is penalized in G-Recall. As a result, models that prioritize ungrounded diversity will experience a significant drop when transitioning from ungrounded recall to G-Recall.

\begin{equation}
\mathrm{G\text{-}Recall} = \frac{1}{|\bar{\mathcal{Q}}|} \sum_{\bar{q} \in \bar{\mathcal{Q}}} V(\bar{q}, \hat{p}),
\label{eq:def_recall}
\end{equation}
We provide further details on how we obtain $\bar{\mathcal{Q}}$ in Appendix~\ref{app:precision_recall}.

\section{Result}

We empirically validate \ours ensures human-level diversity, while balancing with grounding accuracy.
We ablate different components to 
show their contributions and also analyze errors.
\begin{table*}[!ht]
    \centering
    {\fontsize{10pt}{12pt}\selectfont
    \begin{tabular}{l ccc ccc}
        \thickhline
         & \multicolumn{3}{c}{\multirow{1.25}{*}{Existing--Ungrounded}}
         & \multicolumn{3}{c}{\multirow{1.25}{*}{Grounded}}
         \\
        \cmidrule(r){2-4}
        \cmidrule(r){5-7}
        \multirow{-1.25}{*}{Method}
         & \multicolumn{1}{c}{\multirow{-1.25}{*}{$|\hat{\mathcal{Q}}|$}}
         & \multicolumn{1}{c}{\multirow{-1.25}{*}{Sufficient\%}}
         & \multicolumn{1}{c}{\multirow{-1.25}{*}{Recall}}
         & \multicolumn{1}{c}{\multirow{-1.25}{*}{G-Precision}}
         & \multicolumn{1}{c}{\multirow{-1.25}{*}{G-Recall}}
         & \multicolumn{1}{c}{\multirow{-1.25}{*}{G-$\textrm{F}_{\textrm{1}}$}}
        \\\hline
        \texttt{LLaMA 3.1 8B} & & & & & \\
        \phantom{0} DtV~\citep{in-etal-2024-diversify-arxiv} &
        1.36 & \phantom{0}0.21 & 21.36 & 83.72 & \phantom{0}8.14 & 14.84 \\
        \phantom{00} $-$Verification &
            2.26 & \phantom{0}1.37 & \textbf{40.24}
            & 52.75 & 44.43 & 48.23 \\
        \phantom{0} RAC~(\citealp{kim-etal-2023-tree}) &
        0.93 & \phantom{0}8.09 & 11.40 & \textbf{84.27} & 17.43 & 28.89 \\
        \rowcolor{gray!10}
        \phantom{0} \ours (Ours) &
            \textbf{3.78} & \textbf{23.76} & 36.76
            & 61.51 & \textbf{54.77} & \textbf{57.94}
            \\
        \hline

        \texttt{LLaMA 3.3 70B} \\
        \phantom{0} DtV &
        2.06 & \phantom{0}5.60 & 32.44 & 69.89 & 36.55 & 48.00 \\
        \phantom{00} $-$Verification &
            3.68 & 34.73 & \textbf{47.23}
            & 46.25 & 52.16 & 49.03
            \\
        \phantom{0} RAC &
        \textbf{3.78} & \textbf{57.81} & 45.27 & 76.71 & 50.29 & 60.75 \\
        \rowcolor{gray!10}
        \phantom{0} \ours &
            3.70 & 24.21 & 36.66
            & \textbf{81.50} & \textbf{58.04} & \textbf{67.80} 
            \\
        \hline
        
        \texttt{GPT-4o} \\
        \phantom{0} DtV &
        1.73 & \phantom{0}3.80 & 25.57 & 72.34 & 29.38 & 41.79 \\
        \phantom{00} $-$Verification &
            3.07 & 19.51 & \textbf{45.39}
            & 51.35 & 49.85 & 50.59
            \\
        \phantom{0} RAC &
        2.12 & 15.42 & 23.43 & 84.79 & 36.57 & 51.10 \\
        \rowcolor{gray!10}
        \phantom{0} \ours &
            \textbf{3.16} & \textbf{21.49} & 37.41
            & \textbf{92.82} & \textbf{57.25} & \textbf{70.82} 
            \\
        \hline
        
        Human interpretations $\tilde{\mathcal{Q}}$ &
            3.36 & 22.02 & $\cdot$ 
            & 65.47 & $\cdot$ & $\cdot$
            \\
        \thickhline
    \end{tabular}
    }
    \caption{
    Evaluation of diversity and correctness of generated interpretations on ASQA development set, with both ungrounded and grounded metrics.
    `DtV$-$verification' refers to Eq.~\ref{eq:diva_q_extract}.
    }
    \label{tab:main_orig_query}
\end{table*}

\subsection{Experimental Settings}

\subsubsection{Evaluation Metrics}\label{sec:met}

To evaluate diversity,
we report the average number of generated interpretations per query $|\hat{\mathcal{Q}}|$,
the proportion of queries with 
sufficient diversity, denoted as Sufficient\%\footnote{We deduplicate the generated interpretations using the prompt $I_{\textrm{D}}$ in Figure~\ref{fig:prompt_dedup} and count how many unique interpretations are found.}
and (ungrounded) recall, all computed against human interpretations without grounding.

For grounded metrics, we use
our agentic evaluation protocol described in Section~\ref{subsec:eval} to compute G-precision, G-recall, and G-$\textrm{F}_1$ score defined as the harmonic mean of the precision and recall, balancing the objective of two.

\subsubsection{Evaluation Datasets}
We consider the ASQA benchmark~\citep{stelmakh-etal-2022-asqa} as our main target dataset for evaluation.
It is built upon the ambiguous question $q$ from the AmbigNQ benchmark~\citep{min-etal-2020-ambigqa}, adding: (1) 2-6 human-annotated interpretations $\tilde{\mathcal{Q}}$, 
along with corresponding answers $\tilde{\mathcal{Y}}$,
 (2)  optionally attached with a supporting passage $\tilde{p}_i$ per each interpretation $\tilde{q}_i$ from the corpus $C$, Wikipedia.
We utilize its validation split, which consists of 948 ambiguous questions with list of human-annotated interpretations attached to each question, list of answers to those questions, and also long-form answers composed by aggregating those answers.

\subsubsection{Implementation Details}
\label{subsubsec:implt_details}

The retrieval system comprises \texttt{arctic-embed} \citep{merrick-etal-2024-arctic-arxiv}\footnote{\texttt{Snowflake/snowflake-arctic-embed-m-v2.0}} based first-phase retrieval and second-phase passage reranking using \texttt{gte-Qwen2}~\citep{li-etal-2023-towards-arxiv}.\footnote{\texttt{Alibaba-NLP/gte-Qwen2-7B-instruct}}

For the clustering algorithm for consolidation phase described in Section~\ref{subsec:clustering}, we used HDBSCAN~\citep{campello-etal-2013-density-pakdd}, a hierarchical density-based clustering algorithm.
As the encoder $f$ for obtaining embeddings, we reused the same \texttt{gte-Qwen2} embedding model used in retrieval.
For decoding, greedy decoding was used to obtain deterministic responses from LLMs.

For verification in DtV, we fixed the verifier as the most expensive model, GPT-4o, as less capable 8B and 70B models struggled to provide reliable results for long-context inputs, damaging its performance severely.
More details can be found in Appendix~\ref{app:impl_details}.

\subsubsection{Baselines}
\label{subsubsec:baselines}

\begin{table}[t]
\centering
\fontsize{10.5pt}{12pt}\selectfont
\begin{tabular}{lcc}
    \thickhline
    & \phantom{1}(\romannumeral 1)\phantom{1} & (\romannumeral 2) \\ \hline
    DtV & {\textcolor{red}{\ding{55}}} & \ding{51} \\
    RAC~\citep{kim-etal-2023-tree} & {\ding{51}} & {\textcolor{red}{\ding{55}}} \\
    \rowcolor{gray!10}
    \ours (Ours) & {\ding{51}} & {\ding{51}} \\
    \thickhline
\end{tabular}
\caption{
Comparison of baslines, based on (\romannumeral 1) whether diversification is retriever-augmented and (\romannumeral 2) whether joint verification of triple $(\hat{q},\hat{p},\hat{y})$ is involved. 
}
\label{tab:comp_strategies}
\end{table}

While we mainly compare \ours against DtV and its ablated version as baseline,
we also consider retrieval-augmented clarification~(RAC; \citealp{kim-etal-2023-tree})
as contrasted in Table~\ref{tab:comp_strategies}
which leverages relevance feedback
in diversification similar to \ours:
\begin{equation}
\texttt{LLM}(q, U_q; I_{\textrm{G}}).
\label{eq:rac}
\end{equation}

Our distinction is
first extracting
$(\hat{q},\hat{y})$
then jointly verifying relevance and answerability,
while RAC evaluates the entire $U_q$ with a single long-context LLM inference---negative impact of such decision is empirically discussed in Section~\ref{sec:results}.

\subsection{Results}\label{sec:results}

In this section, we validate the effectiveness and efficiency of \ours in disambiguating ambiguous questions, answering the following research questions:
\begin{itemize}
\item RQ1: Is \ours diverse?
\item RQ2: Does diversity balance with grounding?

\item RQ3: What is the effect of consolidation phase, and how is it tunable?
\end{itemize}

\paragraph{\ours ensures human-level diversity.}
Table~\ref{tab:main_orig_query} reports
scores of \ours, human-annotation, RAC~\citep{kim-etal-2023-tree}, and DtV~\citep{in-etal-2024-diversify-arxiv},
in terms of diversity metrics
`$|\hat{\mathcal{Q}}|$',
`Sufficient\%',
and `Recall' (Section~\ref{sec:met}).

Note, in all metrics,
\ours's output is comparable to human interpretation, and remains consistent across diverse models sizes.
In contrast, diversity of interpretations from DtV and RAC tend to be highly affected by model choice/size.
For example, with smaller model, 8B, the output becomes noticeably less diverse, e.g., Sufficient\% is dropped to <1\%.
This validates \ours generalizes better for smaller models and performs consistently over model sizes, by
decomposing context size to a single passage and verifying each $\hat{y}$ at a time, and allowing separate reasoning on each passage.

While ungrounded recall scores are generally higher for DtV, 
we show the negative consequences of  
incentivizing ungrounded diversity in the next section.

\paragraph{\ours ensures verifiable diversity.}

Table~\ref{tab:main_orig_query} shows ungrounded metrics fail to ground to corpus:
35\% of human-annoated 
fall into such category, 
which will incentivize recall when matches are found from generated interpretations.

It also contrasts
ungrounded metrics, with verifiable metrics G-precision, G-recall and G-$\textrm{F}_1$, to show distinctions.
\ours achieves the highest G-precision, of 93\% with GPT-4o and 81\% with LLaMA 70B as the backbone LLM.

These results reinforce our earlier discussion in Section 4.3:  DtV without verification, despite achieving higher ungrounded recall, suffers from the lowest G-precision across all backbone LLMs, highlighting its tendency to generate unverified interpretations. In contrast, both RAC and \ours produce more accurately grounded interpretations, with \ours further benefiting from both retriever and generator feedback. The G-recall results confirm this advantage, showing that interpretations $\hat{q}_i$ generated by \ours are well-supported by their corresponding passages $\hat{p}_i$, ultimately boosting grounded recall to approximately 57\%.

\begin{table}[t]
    \centering
    \fontsize{10pt}{12pt}\selectfont
    \begin{tabular}{lccc}
        \thickhline
        Clustering & $|\hat{\mathcal{Q}}|$ & G-Precision & G-Recall \\ \hline
        \multicolumn{4}{l}{Parameter} \\
        \rowcolor{gray!10}
        \phantom{0} Default & \bf 3.70 & 81.50 & \bf 58.04 \\
        \phantom{0} Conservative & 2.41 & \bf 82.40 & 50.72 \\ \hline
        \multicolumn{4}{l}{Embedding} \\
        \rowcolor{gray!10}
        \phantom{0} $f(\hat{q};\hat{y})$ & \bf 3.70 & \bf 81.50 & \bf 58.04 \\
        \phantom{0}  $f(\hat{q})$ & 3.65 & 78.21 & 57.27 \\
        \thickhline
    \end{tabular}

    \caption{Clustering strategies and performance of \ours with 70B generator. Shaded results are based on default setting, and were copied from Table~\ref{tab:main_orig_query}.
    }
    \label{tab:cluster_strats}
\end{table}

Finally, to balance the conflicting optimization objective of precision and recall, G-$\textrm{F}_1$ is reported.
G-$\textrm{F}_1$ scores show clear gap in performance between \ours and baselines for the goal of pursuing both accurately grounded and diverse interpretations.
For instance, while G-precision of RAC is higher with the 8B backbone, it fails to ensure enough diversity, where the average number of interpretations obtained per question is less than 1.
This leads to significantly lower G-recall and thus G-$\textrm{F}_1$ score, reflecting its failure to provide meaningful disambiguations of the original ambiguous question.

\paragraph{Consolidation with Clustering.}
To continue with our RQ3, Table~\ref{tab:cluster_strats} (top half) shows the consolidation step in \ours can be parameterized
to control the number of resulting interpretations.
Specifically, it can be adjusted to %
favor less clusters, denoted as `conservative.'
DtV approaches do not provide such knobs to control,
as deciding the number of resulting interpretations is up to the LLM.

Table~\ref{tab:cluster_strats} (bottom half) also shows \ours can 
control how to embed
a clustering space.
By default, \ours concatenates interpretation $\hat{q}$ and answer $\hat{y}$, then encodes it with an encoder $f$ to obtain $f(\hat{q};\hat{y})$, utilizing match between answers or the accordance of question-answer.
It is shown that embedding questions only also achieves comparable performance,
suggesting interpretations from \ours are already reliable,
as \ours already pruned much of the noisy questions and passages during diversification.

\begin{figure}[tbp]
\centering
\includegraphics[width=\linewidth]{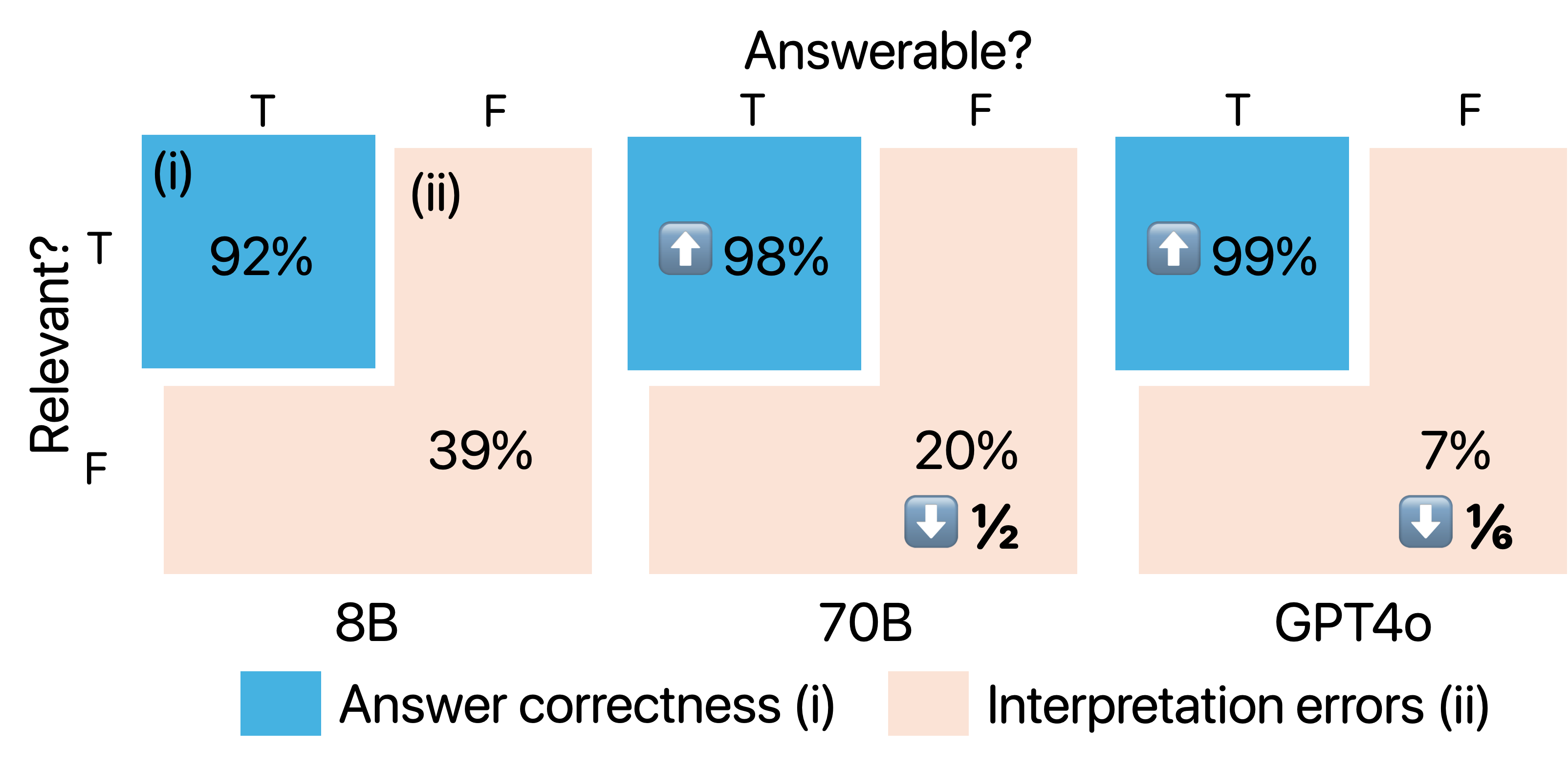}
\caption{
Analysis on accuracy of generated $\hat{q}$'s and $\hat{y}$'s from \ours.
(\romannumeral 1, answer correctness) Models easily derive correct $\hat{q}, \hat{y}$ given an answerable passage. (\romannumeral 2, interpretation error rate) Impact of model scale is more critical in discerning unanswerable passages.
}
\label{fig:error_types}
\end{figure}

\paragraph{Error Analysis.}
We conclude our discussion with error analysis:
Figure~\ref{fig:error_types} shows categorization $(\hat{q}, \hat{y})$ pairs generated from \ours.
Specifically, an LLM judge decided whether $\hat{q}$ is relevant to the original question $q$, and whether $\hat{q}$ can be answered with $\hat{p}$.

First, the top-left corner of each plot in Figure~\ref{fig:error_types} illustrates cases where the retrieved passage is both relevant and answerable. The 8B model correctly answers 92\% of these instances, while stronger models achieve even higher accuracy, reaching 98\% and 99\%, respectively.

Second, the remaining L-shaped sections depict scenarios where the passage $\hat{p}$ is either irrelevant or unanswerable.
These numbers indicate the proportion of incorrect answers due to ungrounded interpretations, showing smaller models struggling more.
However, these errors are well mitigated with scaling model size,
where the 70B model cut errors by more than half and GPT-4o reduces them to less than one-sixth, compared to 8B.

\begin{figure*}[!ht]
\centering
\includegraphics[width=.85\linewidth]{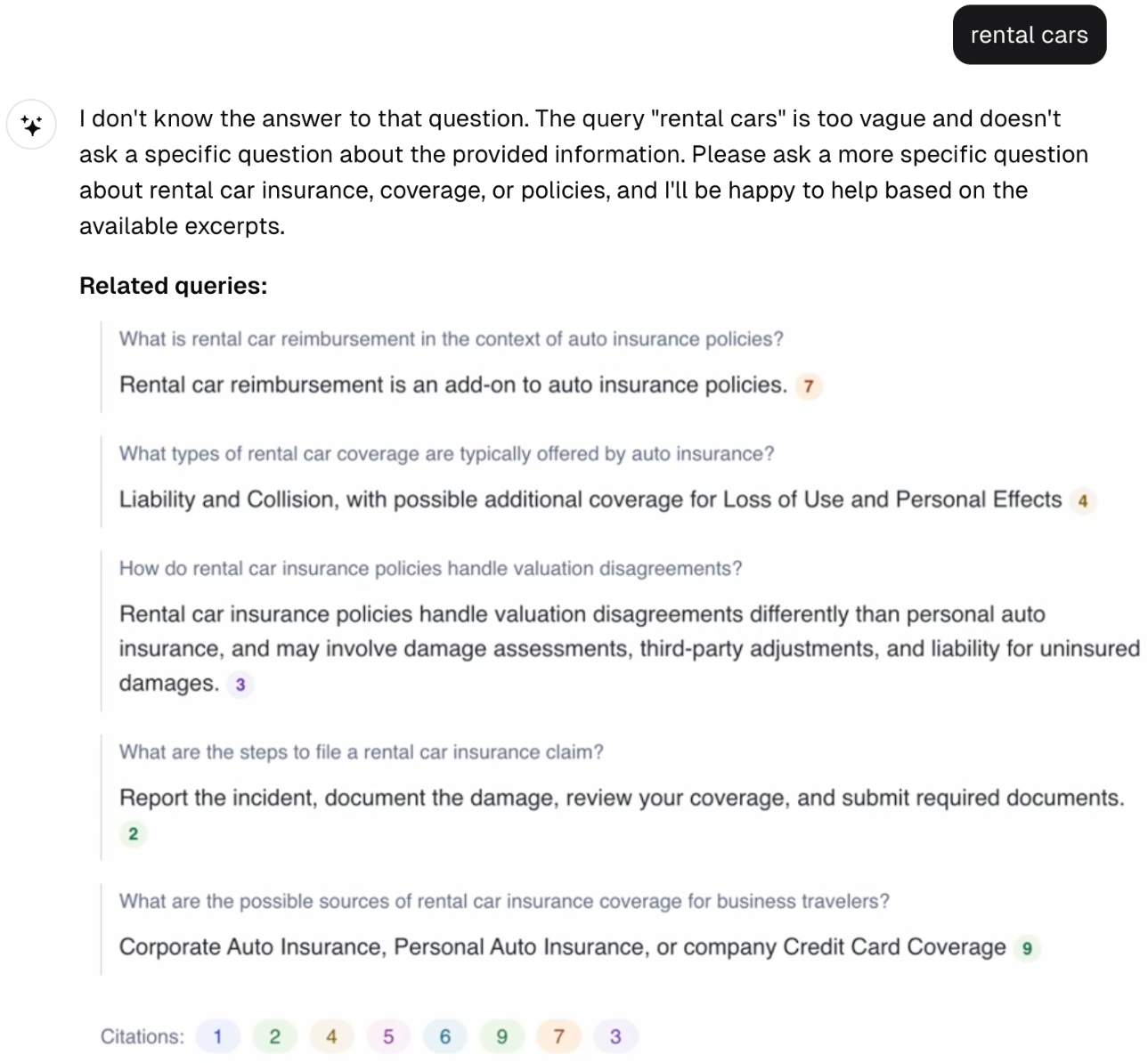}
\caption{Query clarification in Snowflake’s Cortex Agents API setup with tool access to a series of synthetically generated insurance documents retrieved via Cortex Search services.}
\label{fig:arctic}
\end{figure*}

\section{Application in Production }

Figure~\ref{fig:arctic} illustrates a real-world deployment of our method within Snowflake’s Cortex Agents API, highlighting its practical impact. In this production setting, our approach is applied to a synthetic corpus for an insurance company analyzing repair invoices related to auto insurance claims.

The figure demonstrates how a vague inquiry, such as “rental cars,” is disambiguated into grounded clarifications. Each clarified question is grounded to the  cited passages
that can answer the question, obtained from the verification process (Eq. 5). This approach significantly enhances retrieval accuracy and improves the overall user experience.

\section{Conclusion}

We question the common workflow of DtV for grounding ambiguous questions in RAG.
We proposed \ours,
which more efficiently and effectively
grounds diversification to the retrieval corpus, ensuring verifiability of the generated interpretations and answers.
Both ungrounded and grounded evaluation empirically validated our method significantly improves groundedness, while balancing with human-level diversity.

\section*{Limitations}

Though we propose consolidation to handle noisy feedback, we have not considered extreme cases  where the retriever or generator performs unreasonably poor or adversarially.

This would impact the initial retrieval step, where an adaptive scheme can mitigate by better balancing the retrieval breadth (with additional calls) and efficiency.
Developing an approach that allows the agent to dynamically adapt, based on feedback quality, remains an open challenge for future work.

\bibliography{anthology_part1,anthology_part2,anthology_part3,custom}

\clearpage

\appendix

\section{Grounding Evaluation}
\label{app:precision_recall}

Here, we explain in more detail how our evaluation protocol identified matching pairs for computing precision and recall.
We begin by reviewing the setting of previous works and how original precision/recall has been computed in ambiguous question answering evaluation.

\paragraph{Ungrounded Precision/Recall} To obtain precision, the ratio of correct interpretations among generated, correctness labels have been determined by regarding human interpretations $\tilde{\mathcal{Q}}$ as ground-truth.
If the generated interpretation matches one of the human interpretations, it is considered correct; otherwise, it is considered incorrect.
This can be formulated as
\begin{equation}
\mathrm{Precision} = \frac{1}{|\hat{\mathcal{Q}}|} \sum_{\hat{q}\in \hat{\mathcal{Q}}} \texttt{Match} \left(\hat{q},\tilde{\mathcal{Q}}\right), 
\label{eq:def_precision_old}
\end{equation}
where $\texttt{Match}(\cdot, \cdot)$ denotes whether two questions are matched, or, the extended match
\begin{equation}
\texttt{Match}(\hat{q}, \tilde{\mathcal{Q}}) = \mathds{1} \left( \exists_{\tilde{q}\in\tilde{\mathcal{Q}}} \ \texttt{Match}\left(\hat{q},\tilde{q}\right) \right),
\end{equation}
compared against the set of interpretations $\tilde{\mathcal{Q}}$. %
Following the notation from the main text, from this point we denote such matching function as $V$, for the sake of notational simplicity.

Before LLM-as-a-judge was widely adopted, 
$V$ was often instantiated with measures such as lexical-overlap based scores, such as BLEU, exceeding some threshold $\tau$ or not,
\begin{equation}
V \left(\hat{q},\tilde{q}\right) = \mathds{1} \left( \texttt{BLEU}\left(\hat{q},\tilde{q}\right) > \tau \right).
\end{equation}
While such pairwise match can be easily replaced with querying an LLM judge, it would incur $|\hat{\mathcal{Q}}|\times|\tilde{\mathcal{Q}}|$ LLM calls;
for the sake of efficiency, we let the LLM judge to directly determine the match against the set of human interpretations as follows:
\begin{equation}
V\left(\hat{q},\tilde{\mathcal{Q}}\right) = \texttt{LLM}\left(\hat{q},\tilde{\mathcal{Q}}; I_{\textrm{M}}\right).
\label{eq:list_match_llm}
\end{equation}

Similarly, recall, the proportion of ground-truth interpretations successfully generated, is defined as
\begin{equation}
\mathrm{Recall} = \frac{1}{|\tilde{\mathcal{Q}}|} \sum_{\tilde{q}\in \tilde{\mathcal{Q}}}
V \left(\tilde{q},\hat{\mathcal{Q}}\right), 
\label{eq:def_recall_old}
\end{equation}
where whether each $\tilde{q}$ has been covered or not is determined with an LLM in the same way as Eq.~\ref{eq:list_match_llm}.

\paragraph{Grounded Precision/Recall} Our grounded evaluation essentially replaces 
$V(\cdot, \cdot)$ with new matching mechanism that also counts grounding.
For grounding the precision metric, we directly verify if the supporting passage $\hat{p}$ provided along with $\hat{q}$ can answer $\hat{q}$. 
Thus, Eq.~\ref{eq:def_precision_old} is rewritten to consider the `match' between $\hat{q}$ and $\hat{p}$ as
\begin{equation}
\mathrm{Grounded\ Precision} = \frac{1}{|\hat{\mathcal{Q}}|} \sum_{\hat{q}\in \hat{\mathcal{Q}}}
V \left(\hat{q},\hat{p}\right),
\label{eq:def_precision_transition}
\end{equation}
where we replace $\hat{p}$ with
retrieved passages from $U_q$,
proxy to the whole passages $C$,
if such supporting passage $\hat{p}$ is not available, for example for the case of pseudo-interpretations which are obtained independently of retrieval.
Implementation-wise, $V$ is realized with an LLM judge as
\begin{equation}
V(\hat{q}, \hat{p}) = \texttt{LLM}\left(\hat{q},\hat{p};I_{\textrm{V}}\right).
\end{equation}

With the same rationale, we redefine recall as a grounded metric
\begin{multline}
\mathrm{Grounded\ Recall} =
\frac{1}{|\bar{\mathcal{Q}}|} \sum_{\bar{q}\in \bar{\mathcal{Q}}}
V \left( \bar{q}, \hat{p} \right),
\label{eq:def_recall_transition}
\end{multline}
where we also accommodate models that do not provide $\hat{p}$ for each $\hat{q}$ with $\tilde{\mathcal{P}}$, as for precision.

In Eq.~\ref{eq:def_recall_old}, the ground-truth interpretation has been approximated by human interpretations $\tilde{\mathcal{Q}}$.
To increase the recall of this proxy,
we add verified model prediction $V(\hat{q},\hat{p})=1$.
\begin{equation}
\bar{\mathcal{Q}} =  \left\{ \hat{q}\in (\hat{\mathcal{Q}} \cup \tilde{\mathcal{Q}}) \,\middle|\, 
V \left(\hat{q},\hat{p}\right)\right\}.
\end{equation}
Eq.~\ref{eq:def_recall_transition} shows how
$\bar{\mathcal{Q}}$ complements the human interpretations $\tilde{\mathcal{Q}}$.

The prompts used to instruct the judge LLM, $I_\textrm{M}$ for finding a match in a list of questions and $I_\textrm{V}$ for verifying a passage can be found in Appendix~\ref{app:prompts}.

\begin{figure*}[tbp]
\centering
\includegraphics[width=.85\linewidth]{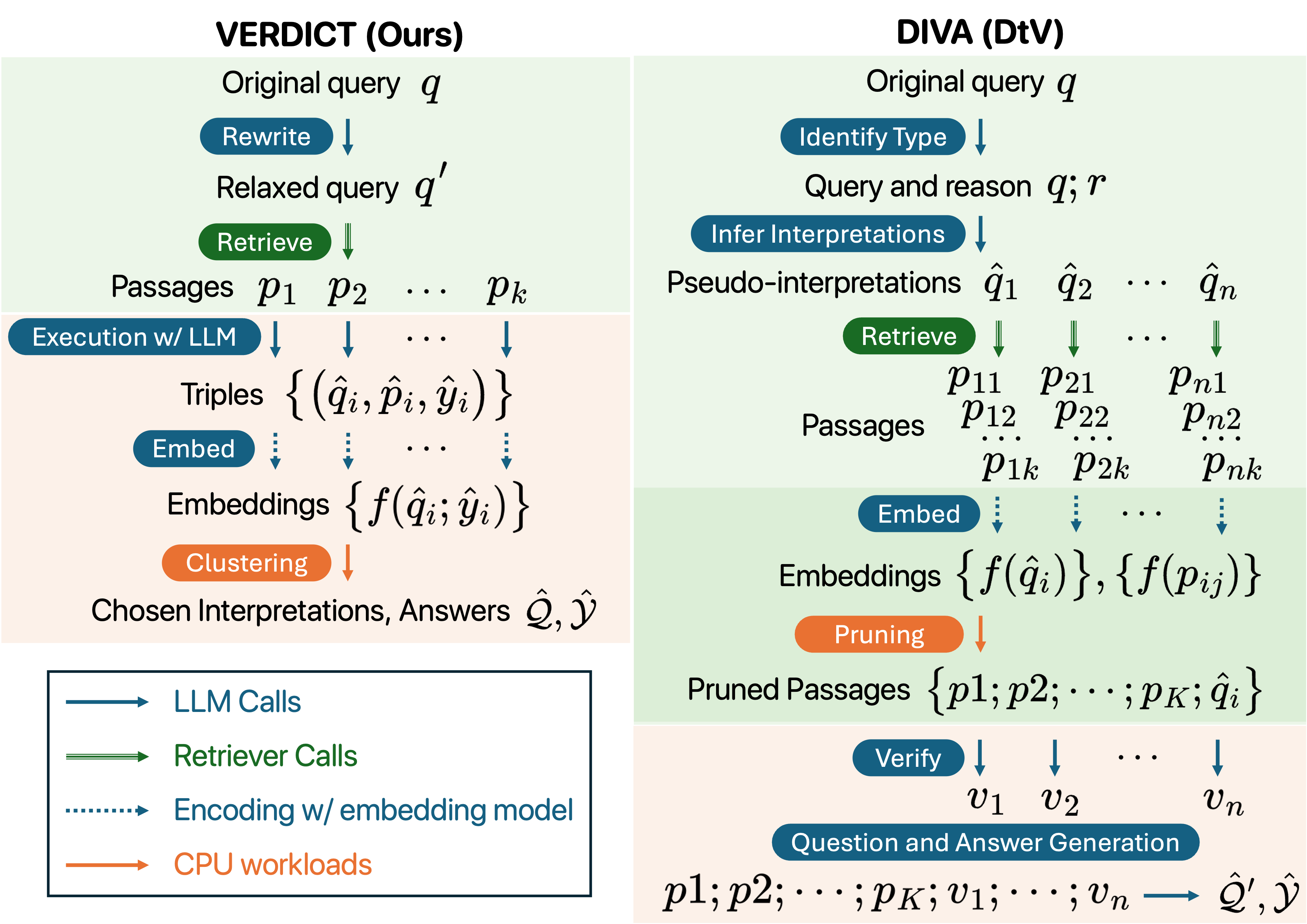}
\caption{
Comparison of end-to-end workflow of \ours and DtV (DIVA; \citealp{in-etal-2024-diversify-arxiv}) for handling ambiguous question answering task. Vertical arrangement denotes sequential dependency, while calls that can run in parallel are placed at the same horizontal level.
}
\label{fig:e2e_compare}
\end{figure*}

\section{E2E Pipeline of \ours and DtV}
\label{app:detailed_workflow}
Figure~\ref{fig:e2e_compare} describes how an ambiguous question is answered, with our \ours and DIVA~\citep{in-etal-2024-diversify-arxiv}, a representative of Diversify-then-Verify (DtV) workflow.
DtV involves more sequential steps, verifying the relevance of passages post hoc while in \ours verification is integrated into diversification, as shown on the left.

\section{LLM Prompts}
\label{app:prompts}
In this section, we provide the instruction prompts provided to LLMs.
For reproducing baselines, we reused their prompts, $I_{\textrm{P}}$ for generating pseudo-interpretations~\citep{in-etal-2024-diversify-arxiv} or $I_{\textrm{G}}$ for generating a list of interpretations and answers at once~\citep{kim-etal-2023-tree}, which can be found in their respective papers.

Figure~\ref{fig:prompt_q_extract} shows the prompt used to generate interpretation and answer in \ours, as described in Eq.~\ref{eq:ours_q_extract_single}.
Figure~\ref{fig:prompt_q_Q_match} and \ref{fig:prompt_q_p_veirfy} present prompts used for evaluation, as discussed in detail in Appendix~\ref{app:precision_recall}.

\begin{figure*}[!htbp]
\centering
\begin{tcolorbox}[boxrule=0pt, title=Prompt for Verified Diversification]
Given an ambiguous query and one of the passages from retrieval results, provide a disambiguated query which can be answered by the passage.
Try to infer the user's intent with the ambiguous query and think of possible concrete, non-ambiguous rewritten questions.
If you cannot find any of them, which can be answered by the provided document, simply abstain by replying with `null'.
You should provide at most one subquestion, the most relevant one you can think of.\\

Here are the rules to follow when generating the question and answer:\\
1. The generated question must be a disambiguation of the original ambiguous query.\\
2. The question should be fully answerable from information present in given passage. Even if the passage is relevant to the original ambiguous query, if it is not self-contained, abstain by responding with `null'.\\
3. Make sure the question is clear and unambiguous, while clarifying the intent of the original ambiguous question.\\
4. Phrases like `based on the provided context', `according to the passage', etc., are not allowed to appear in the question. Similarly, questions such as ``What is not mentioned about something in the passage?'' are not acceptable.\\
5. When addressing questions tied to a specific moment, provide the clearest possible time reference. Avoid ambiguous questions such as ``Which country has won the most recent World Cup?'' since the answer varies depending on when the question is asked.\\
6. The answer must be specifically based on the information provided in the passage. Your prior knowledge should not intervene in answering the identified clarification question.\\

Input fields are:\\
\textbf{Question}: \{\texttt{ambiguous question} ($q$)\}\\
\textbf{Passage}: \{\texttt{passage} ($p$)\}\\

Output fields are:\\
\textbf{Interpretation}:  \{\texttt{generated interpretation} ($\hat{q}$)\} \\
\textbf{Answer}:  \{\texttt{generated answer} ($\hat{y}$)\}
\end{tcolorbox}
\caption{Prompt $I_{\textrm{E}}$ for obtaining interpretation $\hat{q}$ and answer $\hat{y}$ with execution feedback from the LLM.}
\label{fig:prompt_q_extract}
\end{figure*}

\begin{figure*}[!htbp]
\centering
\begin{tcolorbox}[boxrule=0pt, title=Evaluation Prompt for Ungrounded Precision/Recall]
Given a list of generated disambiguated subquestions that clarify the intent of an ambiguous question, compare them with the list of predefined subquestions and determine how many have been successfully identified. You should return a binary label, Yes or No, for each subquestion indicating whether it was covered or not.\\

Input fields are:\\
\textbf{Question}: \{\texttt{ambiguous question} ($q$)\}\\
\textbf{Generated Disambiguations}: \{\texttt{generated interpretations} ($\hat{\mathcal{Q}}$)\}\\
\textbf{Ground-truth Disambiguations}: \{\texttt{human-annotated interpretations} ($\tilde{\mathcal{Q}}$)\}\\

Output fields are:\\
\textbf{Decisions}:  \{\texttt{match} ($V(\hat{q},\tilde{q})$'s)\} \\
\end{tcolorbox}
\caption{Prompt $I_{\textrm{M}}$ for determining matches between $\hat{q}$'s and $\tilde{q}$'s.}
\label{fig:prompt_q_Q_match}
\end{figure*}

\begin{figure*}[!htbp]
\centering
\begin{tcolorbox}[boxrule=0pt, title=Evaluation Prompt for Verification]
Given a question, an answer and an associated passage, decide if the passage can support the answer, providing enough evidence to reach the answer given the question.
Your answer should be either Yes or No.\\

Input fields are:\\
\textbf{Question}: \{\texttt{interpretation} ($\hat{q}$)\}\\
\textbf{Passage}: \{\texttt{passage} ($\hat{p}$)\}\\

Output fields are:\\
\textbf{Decision}:  \{\texttt{match} ($V(\hat{q},\hat{p})$)\} \\
\end{tcolorbox}
\caption{Prompt $I_{\textrm{V}}$ for determining a match between $\hat{q}$ and $\hat{p}$.}
\label{fig:prompt_q_p_veirfy}
\end{figure*}

\begin{figure*}[!htbp]
\centering
\begin{tcolorbox}[boxrule=0pt, title=Evaluation Prompt for Deduplication]
Given a list of subquestions, which are derived disambiguations of an ambiguous query,
remove nearly identical duplicates and leave only distinct ones.
You should provide a list of the remaining subquestions, one at a line.\\

Input fields are:\\
\textbf{Ambiguous Question}: \{\texttt{ambiguous question} ($q$)\}\\
\textbf{List of Disambiguated Subquestions}: \{\texttt{interpretations} ($\hat{\mathcal{Q}}$ or $\tilde{\mathcal{Q}}$)\}\\

Output fields are:\\
\textbf{List of Unique Subquestions}:  \{\texttt{deduplicated interpretations} \} \\
\end{tcolorbox}
\caption{Prompt $I_{\textrm{D}}$ for removing (near-)duplicates in a list of interpretations for an ambiguous question.}
\label{fig:prompt_dedup}
\end{figure*}

\section{More on Implementation Details}
\label{app:impl_details}

Here, we provide more details regarding the experimental settings.
For main experiments, \ours (ours) and RAC have used top-20 passages, retrieved with relaxed question $q'$ as the search query.
For DtV, we have used the same encoder, Qwen, for pruning the passages and then fixed verifier LLM as GPT-4o, as mentioned previously in Section~\ref{subsubsec:implt_details}.
Following the original paper's setting, we retrieved top-5 passages for each pseudo-interpretation, then finally selected top-5 among them, after pruning.

\end{document}